\documentclass[final,5p]{elsarticle}

\usepackage{amssymb}

\usepackage[pagewise]{lineno}

\usepackage{bm}
\usepackage{amsfonts}

\usepackage{float}
\usepackage{graphicx}
\usepackage{setspace}
\usepackage{color}
\usepackage{amsmath}
\usepackage{wrapfig}

\graphicspath{{./pdf/}{./pic/}}
\usepackage{ifpdf}
\ifpdf   
  \graphicspath{{./pdf/}{./figures/}}
  \DeclareGraphicsExtensions{.pdf,.png,.jpg}
\else    
  \graphicspath{{./eps/}}
  \DeclareGraphicsExtensions{.eps,.ps}
\fi

\usepackage{multirow}
\usepackage{booktabs} 
\usepackage{algorithmic,algorithm}

\usepackage{hyperref}
\hypersetup{bookmarks=true,
    bookmarksdepth=3,
    bookmarksopen,
    bookmarksnumbered,
    pdfstartview=FitH,
    colorlinks=true,
    breaklinks=true,
}

\usepackage{dcolumn}

\newcolumntype{I}{!{\vrule width 0.6pt}}

\usepackage{color}

\journal{Pattern Recognition}

\begin{document}

\begin{frontmatter}



\title{Instance-Aware Representation Learning and Association\\for Online Multi-Person Tracking}


\author[label1,label2,label6]{Hefeng Wu}
\ead{wuhefeng@gmail.com}
\author[label3]{Yafei Hu}
\author[label4]{Keze Wang}
\author[label5]{Hanhui Li}
\author[label1]{Lin Nie\corref{cor1}}
\ead{nielin@mail.sysu.edu.cn}
\author[label1]{Hui Cheng}

\address[label1]{Sun Yat-sen University, Guangzhou 510006, China}
\address[label2]{Guangdong University of Foreign Studies, Guangzhou 510006, China}
\address[label3]{Carnegie Mellon University, Pittsburgh, PA 15213, USA}
\address[label4]{University of California, Los Angeles, CA 90024, USA}
\address[label5]{Guilin University of Electronic Technology, Guilin 541004, China}
\address[label6]{WINNER Technology, China}
\cortext[cor1]{Corresponding author is Lin Nie.}

\begin{abstract}
Multi-Person Tracking (MPT) is often addressed within the detection-to-association paradigm. In such approaches, human detections are first extracted in every frame and person trajectories are then recovered by a procedure of data association (usually offline). However, their performances usually degenerate in presence of detection errors, mutual interactions and occlusions. In this paper, we present a deep learning based MPT approach that learns instance-aware representations of tracked persons and  robustly online infers states of the tracked persons. Specifically, we design a multi-branch neural network (MBN), which predicts the classification confidences and locations of all targets by taking a batch of candidate regions as input. In our MBN architecture, each branch (instance-subnet) corresponds to an individual to be tracked and new branches can be dynamically created for handling newly appearing persons. Then based on the output of MBN, we construct a joint association matrix that represents meaningful states of tracked persons (e.g., being tracked or disappearing from the scene) and solve it by using the efficient Hungarian algorithm. Moreover, we allow the instance-subnets to be updated during tracking by online mining hard examples, accounting to person appearance variations over time. We comprehensively evaluate our framework on a popular MPT benchmark, demonstrating its excellent performance in comparison with recent online MPT methods.
\end{abstract}


\end{frontmatter}


\section{Introduction}

\begin{figure}[!tb]
    \centering
    \includegraphics[width= 0.99 \columnwidth]{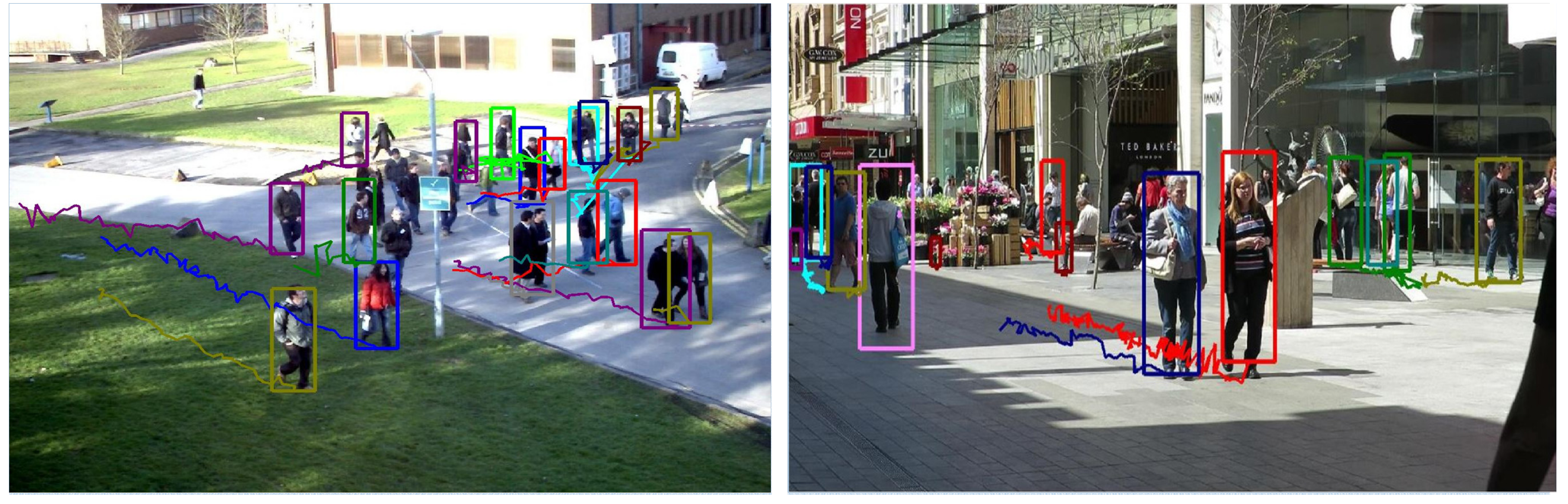}
    \caption{Ambiguities of multi-person tracking arise under complex scenarios such as unknown numbers of targets, mutual interactions, occlusions over time.}
    \label{fig:static}
\end{figure}

Multi-Person Tracking (MPT), as a key component of several intelligent applications such as automatic driving and video surveillance, has attracted special attention beyond general object tracking. The goal of MPT is to estimate the states of multiple observed persons while preserving their identifications under appearance variation over time. Existing MPT methods are mainly developed within the detection-to-association paradigm, where human in each frame are usually detected by pre-trained classifiers and associated for identifying the trajectories of persons throughout video sequences. Recently proposed MPT methods have shown impressive performance improvement thanks to the development of object (pedestrian) detectors (e.g., deep learning based models). Nevertheless, the problem still remains unsolved in complex scenes (see Fig. \ref{fig:static} for examples) due to the following reasons:

\begin{itemize}
  \item Mutual interactions and occlusions of moving persons usually degenerate the performances of human detectors and the resulting false positive detections increase the complexity of conserving person identifications.
  \item It is quite difficult to handle ambiguities caused by person appearance and motion variations throughout sequences. Some offline methods (i.e., by exploiting detections from a span of deferred observations) are usually adopted but not suitable for realistic applications (i.e., working with less observed data).
\end{itemize}

To address the abovementioned issues, in this work we propose to amend the traditional detection-to-association paradigm by learning instance-aware person representations. Unlike the existing methods that usually employ generic (category-level) human detectors, our approach targets on assigning each moving person a specific tracker to reduce ambiguities in complex scenes.
Additionally, modern advances in the development of deep feature representation learning \cite{LinWZF016pami,XieDZWF17pami,WuYL17pr} for object appearance have created new opportunities for MPT methods, which partially motivate us to learn instance-level object representations by deep neural nets.
Therefore, we develop a multi-branch neural network (MBN) that dynamically learns instance-level representations of tracked persons at a low cost, which facilitates robustly online data association for multiple target tracking and thus gives birth to our INstance-Aware Representation Learning and Association (INARLA) framework.

The proposed MBN architecture consists of three main components: i) a shared backbone-net for extracting convolutional features of input regions, ii) a det-pruning-subnet for rejecting the regions from human detection proposals and iii) a variable number of instance-subnets for measuring the confidence of the remaining candidate regions with respect to the tracked targets. Each instance-subnet explicitly corresponds to an individual in the scene and can be online updated by mining hard examples. Moreover,  new instance-subnets can be dynamically created to handle newly appearing targets. In this way, our MBN enables to improve the trackers' robustness by adaptively capturing appearance variations for all the targets over time. Moreover, it is beneficial to relieve the burden of the following step of data association. Traditional detection-to-association trackers usually rely on an expensive step for associating observed data with trajectories (identifications) by establishing spatio-temporal coherence, especially for those offline methods \cite{milan2015joint,wang2015learning}. In contrast, our INARLA framework handles it in a simple and efficient way, thanks to the MBN that can provide powerful instance-level affinity measures for the observed regions. Specifically, we construct a joint association matrix based on the outputs of MBN. This matrix can be divided into four blocks that represent meaningful states of tracked persons (e.g., being tracked or disappearing from the scene), and it results in a standard assignment problem that can be solved efficiently by the Hungarian algorithm \cite{munkres1957algorithms}. In sum, our approach handles the problem of online multi-person tracking with the following steps: i) initializing generic human detections in an input video frame; ii) pruning low-confidence human detections via the det-pruning-subnet; iii) predicting the location of each being tracked individual via its corresponding instance-subnet; iv) inferring the states of all targets by constructing an association matrix with results of step ii) and iii); v) making the MBN network updated according to the inferred states of the targets.

The main contributions of this paper are summarized as follows. First, it presents a novel deep multi-branch neural network that enables dynamically instance-aware representation learning to address realistic challenges in multi-person tracking. Second, it presents a simple yet effective solver for data association based on the deep architecture, which is capable of inferring the states of tracked individuals in a frame-by-frame way. Experimental results on a standard benchmark underline our method's favorable performance in comparison with existing multi-person tracking methods.

\begin{figure*}[!htb]
    \centering
    \includegraphics[width=0.95 \textwidth]{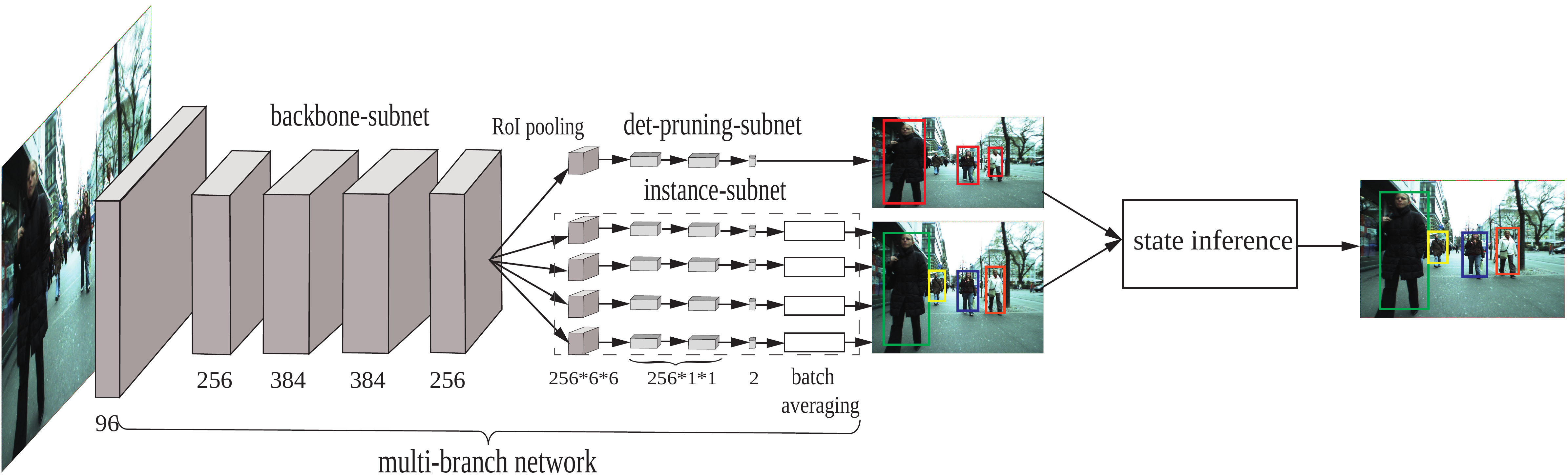}
    \caption{Illustration of our INARLA framework for multi-person tracking. The left side is the architecture of the MBN network. The topmost branch (det-pruning-subnet) excludes false person detections, while the other branches (instance-subnets) predict their corresponding targets independently. Based on the outputs of MBN, we propose an efficient algorithm to jointly infer the state of each person. Best viewed in colour.}\label{fig_network}

\end{figure*}

\section{Related Work}

In literature much efforts have been dedicated in multi-object tracking (MOT), and we review them according to their main technical components, i.e., object representation and data association.

\subsection{Object representation}
How to represent objects plays an important role in MOT for affinity computation or linking object detections across frames.
Many different cues have been presented in the literature, e.g., appearance, location and motion.

Earlier MOT works mostly adopt hand-crafted features for object representation \cite{ChoiS10eccv,DalalT05cvpr,Andriyenko2011Multi,QianYG13pr}. Color histograms are commonly used to represent object appearance in multi-object tracking \cite{ChoiS10eccv,Bae2014Robust}, and histograms of oriented gradients (HOG) \cite{DalalT05cvpr} is also a popular choice \cite{BenfoldR11cvpr,LiWZLLW17tip}. In \cite{Andriyenko2011Multi}, optical flow that reflects the motion information is incorporated for object representation.
In addition, appropriate fusion of multiple cues can yield improved results \cite{KimKFH12accv,lenz2015followme,WuGCW18nca}. Moreover, sophisticated machine learning techniques \cite{Bae2014Robust,FelzenszwalbGMR10pami} are introduced to better describe object appearance models.
However, conventional object representation methods are often badly affected by challenging factors like illumination variations, object deformation, background clutters, etc., which limits their performance and generalization ability to various complex scenarios.

Recently, researchers actively learn object appearance features with deep learning based models due to their powerful representation learning ability, e.g., convolutional neural networks (CNNs) \cite{Wang2015Visual,LiWLL18jcam} and recurrent neural networks (RNNs) \cite{Cui2016CVPR,Ondruska2016Deep}.
A fully convolutional neural network is adopted in \cite{Wang2015Visual} for object tracking, where features from top and lower layers that characterize the target from different perspectives are jointly used with a switch mechanism.
In \cite{Cui2016CVPR}, a  recurrently target-attending tracking method is presented, which attempts to identify
and exploit reliable parts that are beneficial for the tracking process.
But these mentioned deep learning based methods mainly focus on single object tracking with the object being indicated at the first video frame.
As for MOT, recently Leal-Taixe et al. \cite{Leal2016CVPRWorkshops} exploit siamese CNN  for pairwise pedestrian similarity mesurement in offline tracking, while Gaidon and Vig \cite{gaidon2015online} take advantage of the convolutional features in online domain adaption between instances and category in a Bayesian tracking framework. Different from these methods, in this paper we employ a MBN network for instance-aware object representations, in which a backbone-subnet is trained with a novel multi-task loss and instance-subnets are dynamically initialized from a det-pruning-subnet and trained discriminatively online.

\subsection{Data association}
To address the data association problem, existing MOT works can mainly be roughly divided into two categories: offline methods \cite{milan2015joint,wang2015learning,lenz2015followme} and online methods \cite{KimKFH12accv,gaidon2015online,yoon2016online}.

Most MOT methods belong to the first category and process the video in an offline way, where the data association is optimized over the whole video or a span of frames and requires future frames to determine objects' states in the current frame.
Network flow-based MOT methods \cite{Tang2017Multiple,WangTFF16pami} are quite typical in this category, and they generally solve the MOT problem using minimum-cost flow optimization. In \cite{Tang2017Multiple}, linking person hypotheses over time is formulated as a minimum cost lifted multicut problem.
In order to track interacting objects well, Wang et al. \cite{WangTFF16pami} propose novel intertwined flows to handle this issue. Integer program is also often used for formulating data association in MOT \cite{LeibeSG07iccv,WangTFF14eccv}.
In \cite{LeibeSG07iccv}, the quadratic integer program formulation is solved to local optimality by custom heuristics based on recursive search. Mixed integer program is introduced to handle the interaction of multiple objects in \cite{WangTFF14eccv}.
In \cite{MaksaiWFF17iccv}, a non-Markovian approach is proposed to impose global consistency by using behavioral patterns
to guide the association.
These offline methods generally yield better performance by incorporating future frames into formulation and optimization, but this characteristic and the resulted high complexity also add great constraints to their application.

The online methods only use information up to the current frame and require no deferred processing, which are more practical in real-world applications.
In \cite{KimKFH12accv}, the data association between consecutive frames is formulated as bipartite matching and solved by structural support vector machines. Bae et al. \cite{Bae2014Robust} perform online multi-object tracking by combination of local and global association based on tracklet confidence.
Recently, more sophisticated learning methods are introduced to handle this problem.
In \cite{xiang2015learning}, the online association is modeled by Markov Decision Process (MDP) with reinforcement learning.
In \cite{MilanRDSR16}, RNNs are employed to learn the data association from data for online multi-object tracking.
While the recent works spend costly computation in online joint association, this paper introduces an efficient solver for the online association based on the outputs of the MBN network.

\section{Instance-Aware Representation Learning}\label{sec_proposed}

Our INARLA framework incorporates instance-aware representation learning into joint association for online multi-person tracking and can combine with any human detector. As shown in Fig. \ref{fig_network}, we train a multi-branch neural network (MBN) for instance-aware representation learning. In a new frame, our approach embeds the MBN network's outputs in an association matrix to jointly infer the objects' states, which will be fed back to the MBN network.

\subsection{Multi-branch neural network}\label{sec:MBN}

The architecture of our MBN network is illustrated in Fig. \ref{fig_network}, which consists of three main components: a shared backbone-subnet, a det-pruning-subnet and a variable number of instance-subnets.
The backbone-subnet is fully convolutional and can take an image of arbitrary size as input to extract convolutional features. Among the branch subnets, the det-pruning-subnet is designed to evaluate and reject the noisy person proposals from a public human detector and also to initialize instance-subnets, while each instance-subnet predicts the location of its tracked person and also outputs the confidence score of a candidate being the target.

We build the MBN network from the Fast R-CNN model \cite{girshick2015fast} using CaffeNet \cite{krizhevsky2012imagenet}. We borrow the lower five layers from Fast R-CNN architecture as our backbone-subnet, while the branch subnet structure is specially defined to accommodate our task.
Different branch subnets have the same structure definition. In order to handle the online learning of tracked instances with few examples, we define a lightweight branch subnet architecture, which comprises a region-of-interest (ROI) layer, and three fully connected layers with size of 256, 256 and 2, respectively.

\subsection{Network learning}\label{sec:network_learn}
For concise description, we use $F_{bb}$ to denote the backbone-subnet and $F_i$ to denote the $i$th branch subnet. The 0th branch is the det-pruning-subnet and the $i$th branch ($i\ge1$) is the $i$th instance-subnet, which dynamically changes in conformance with the number of maintained persons.
In addition, $f_i$ denotes the corresponding network that is formed by subnet $F_{bb}$ and the $i$th branch subnet $F_i$ (i.e. $f_i=F_{bb}+F_i$).

The backbone-subnet $F_{bb}$ is initialized from the Fast R-CNN model trained on the large-scale VOC datasets \cite{girshick2015fast}. We initialize the det-pruning-subnet $F_0$ from zero-mean Gaussian distributions with standard deviation 0.01.

We train the network $f_0=F_{bb}+F_0$ offline, and employ a multi-task loss $\mathcal{L}$ on each labeled RoI to jointly optimize
for classification and distance metric embedding:

\begin{equation}\label{eq:multi}
\mathcal{L} = {L_{cls}}(p,u)  + \mu L(x^{\bot}|x^+,\mathcal{X^-})
\end{equation}
where ${L_{cls}}(p,u) = -\log p_u$ is defined as the log loss function over two classes. $p=(p_0, p_1)$ is computed by a softmax over the 2 outputs in the final fully connected layer, and $u$=1 indicates the target and $u$=0 otherwise.

\begin{figure}[!htb]
    \centering
    \includegraphics[width=0.99 \linewidth]{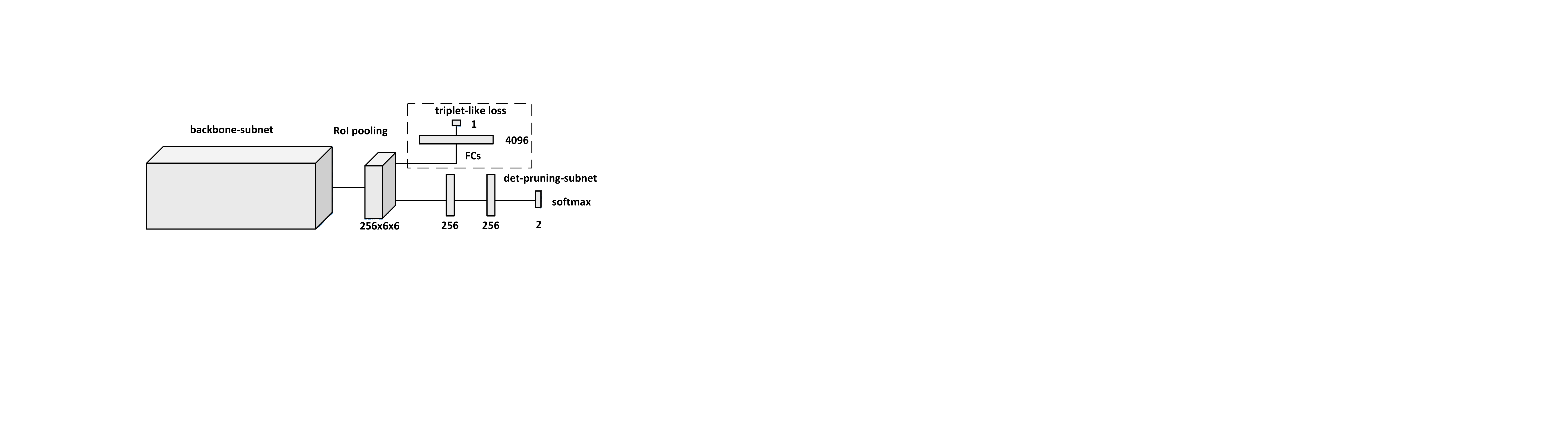}
    \caption{Multi-task learning of the network $f_0=F_{bb}+F_0$.}\label{fig_backbone}
\end{figure}

As illustrated in Fig. \ref{fig_backbone}, we add an auxiliary subnet (in the dashed-line box), consisting of two fully connected layers with sizes 4096 and 1, respectively. A triplet-like loss is used:  $L(x^{\bot}|x^+,\mathcal{X^-})=\sum_{x^-\in\mathcal{X^-}}\phi(D(\mathcal{H}(x^{\bot}),\mathcal{H}(x^-))-D(\mathcal{H}(x^{\bot}),\mathcal{H}(x^+)))$. Here $x^{\bot}$ and $x^+$ are positive examples of the same human object (e.g., sampled nearby or at different frames), while $X^-$ denotes a set of negative examples. $\mathcal{H}(\cdot)$ denotes the 4096-dimensional feature vector, and $D(\cdot, \cdot)$ is the $L_2$ norm distance (i.e. $D(\mathcal{H}(x), \mathcal{H}(y))=\|\mathcal{H}(x)-\mathcal{H}(y)\|_2^2$). The function $\phi(x)$ is defined as $\phi(x)=\log_2(1+2^{-x})$  \cite{YunRV14nips}.

This triplet-like loss can drive similar (dissimilar) examples close to (apart from) each other in the feature space. Optimizing the multi-task loss Eq. (\ref{eq:multi}) can make the feature exacted by the backbone-subnet suitable for discriminating both human/non-human objects and different humans, which is helpful for later instance-subnet training and prediction. To maintain the balance of positive and negative examples, we set the cardinality of $\mathcal{X^-}$ as 2. Thus the batch size for optimization is a multiple of 4. The hyperparameter $\mu$ in Eq. (\ref{eq:multi}) is set as 0.7 in our experiments.

In optimization process, the gradients of the triplet-like loss $L(x^{\bot}|x^+,\mathcal{X^-})$ with respective to the vector $\mathcal{H}(x)$ can be calculated based on the chain rule:

\begin{equation}\label{eq:derivative}
\left\{\begin{array}{ll}
\frac{\partial{L}}{\partial{\mathcal{H}(x^{\bot})}}=2\sum\nolimits_{x^-\in\mathcal{X^-}}\psi_{c}(\mathcal{H}(x^-)-\mathcal{H}(x^+))\\

\frac{\partial{L}}{\partial{\mathcal{H}(x^+)}}=2\sum\nolimits_{x^-\in\mathcal{X^-}}\psi_{c}(\mathcal{H}(x^{\bot})-\mathcal{H}(x^+))\\

\frac{\partial{L}}{\partial{\mathcal{H}(x^-)}}=2\sum\nolimits_{x^-\in\mathcal{X^-}}\psi_{c}(\mathcal{H}(x^-)-\mathcal{H}(x^{\bot}))\\
\end{array}\right.
\end{equation}
where $\psi_{c}=(1+2^{d_{c}})^{-1}$ and $d_c=D(\mathcal{H}(x^{\bot}),\mathcal{H}(x^-))-D(\mathcal{H}(x^{\bot}),\mathcal{H}(x^+))$.

We train the network $f_0$ in a hard-example-mining scheme \cite{Shrivastava2016CVPR}.
Specifically, we start with a dataset of positive examples and a random set of negative examples.
The network $f_0$ is trained to converge on this dataset and subsequently applied to a larger dataset to harvest false positives. Then the network is trained again on the augmented training set with the false positives added. The auxiliary subnet is removed when training is finished.

In the test stage, the instance network $f_i$ ($i\ge1$) is created dynamically by adding a new branch instance-subnet $F_i$ and trained online when a person is newly detected.
The new instance-subnet $F_i$ is initialized from subnet $F_0$, and further trained using only the classification loss ${L_{cls}}(p,u)$ by setting $\mu=0$ in Eq. (\ref{eq:multi}).

We collect $N^+$ (=500) positive samples and $N^-$(=256) negative samples. The intersection-over-union (IoU) overlap ratios of positive and negative samples with this target's detection bounding box are greater than $\theta_1$ (=0.5) and less than $\theta_0$ (=0.3), respectively.
Beyond that, we collect $N^+$ positive samples from every other object as negative samples for this new target to make its specific subnet more discriminative.
In updating, we exploit hard negative examples for online training in the hard-example-mining scheme.
Given a sample $x$, the score $f_i(x)$ measures the similarity between the sample $x$ and the person target $i$.

\subsection{Instance prediction}\label{sec:prediction}
In frame $t$, we apply the proposed MBN network for instance prediction tasks.
An instance-subnet independently predicts the corresponding target's location $x_t$, which consists of center coordinates $(c_x, c_y)$, width $l_w$ and height $l_h$.
We sample $Q$ candidates $\{{s_t^{k}}\}{_{k=1:Q}}$ varying in displacement and scale for each target from its previous location $x_{t-1}$. Specifically, a candidate is denoted as $s_t^k=(c_x+\delta_x,c_y+\delta_y,l_w\cdot\delta_l,l_h\cdot\delta_l)$, with $(\delta_x,\delta_y,\delta_l)$ drawn from a normal distribution whose mean is $(0,0,1)$ and covariance is a diagonal matrix with diagonal vector $\sigma_s$.
The candidates of the target $i$ will pass the network $f_i$ and get their scores {$\{f_i({{s_t^k}})\}$}. Most previous works select the candidate with the maximum score as the optimal location. However, this strategy renders unstable prediction. It is because our features are extracted from a downsampling layer, and candidates with similar locations may be projected to the same region in the feature map and thus get the same feature after RoI pooling. Such instability will be more drastic for small-sized objects. We use a simple and effective scheme to overcome this problem by averaging all the locations whose score over $\alpha\cdot \max_{k=1:Q}f_i({s_t^{k}})$.
So the predicted location of target $i$ will be calculated as

\begin{equation} \label{equ:eq1}
x_{t}^i=mean(\{{s{_t^{k}}}\,|\,f_i({s{_t^{k}}})>\alpha\cdot \max_{k=1:Q}\,f_i({{s_t^{k}}})\} )
\end{equation}

\section{Joint State Inference for Tracking}\label{sec_states}

\begin{figure}[!thb]
    \centering
    \includegraphics[width= 0.99\columnwidth]{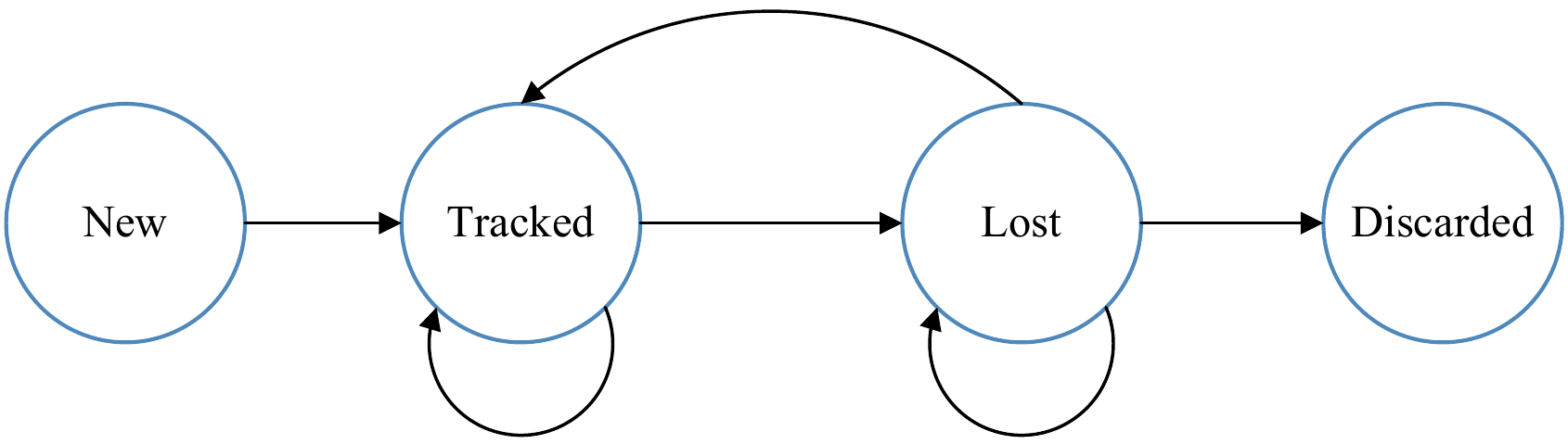}
    \caption{State transition of an individual.}
    \label{fig:lifetime}
\end{figure}

Different states are employed to describe a person target in the video, and Fig. \ref{fig:lifetime} shows the state transition. A person in the ``New'' state denotes being newly detected, and a new identity will be assigned to it (a new instance-subnet will be initialized as well) before it transits to the ``Tracked'' state. When the ``Tracked'' person is considered not found in a frame, its state will be changed to the ``Lost'' state. The ``Lost'' person is still maintained and continues to be looked for, and it will transit to the ``Tracked'' state again if it is found. However, if the ``Lost'' person stays in this state for a certain amount of frames, it will be changed to the ``Discarded'' state, and all its information (identity and instance-subnet) will be removed.
Based on the outputs of MBN, we propose an efficient solver for the joint state inference.

\begin{figure*}[!htb]
    \centering
    \includegraphics[width=1 \textwidth]{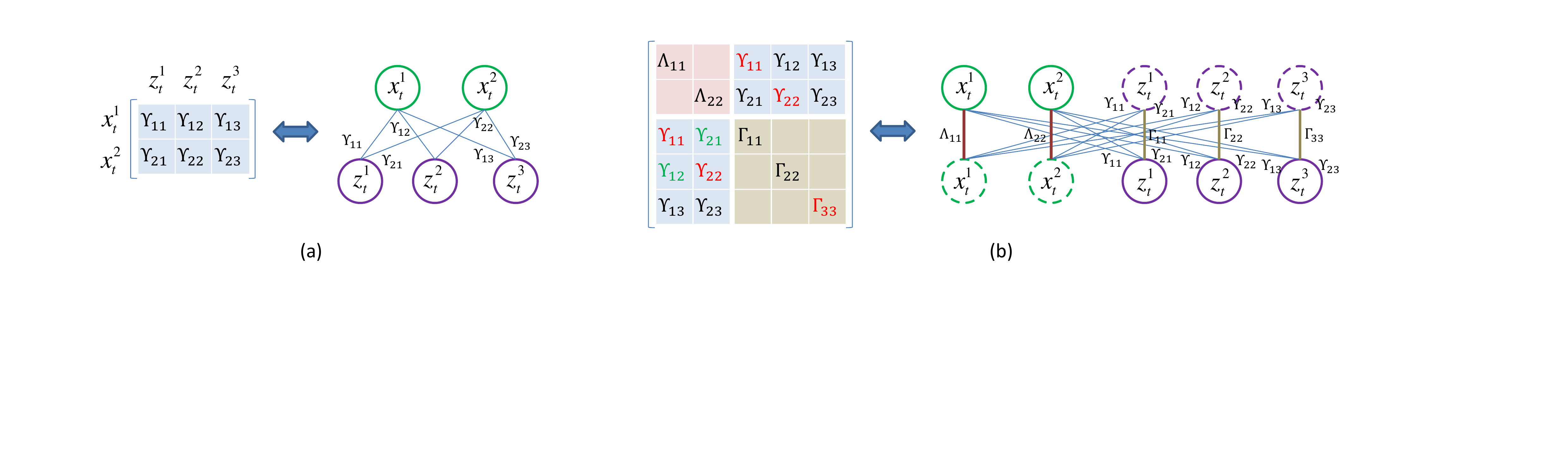}
    \caption{Illustration of joint association matrix. (a) The conventional association matrix and its equivalent bipartite graph. (b) Our joint association matrix and its equivalent graph. $\{x_t^i\}_{i=1}^2$  are instance predictions, $\{z_t^j\}_{j=1}^3$ are person observations, they serve as nodes in the equivalent graph and the matrix elements serve as edge weights. See text for explanations. }\label{fig:association}

\end{figure*}

\subsection{Joint association matrix construction} \label{sec_td}

Assume that we maintain $M$ tracked person targets and there exist $N$ new person observations in frame $t$ after applying the proposed MBN network.
Let $\{x_t^i\}_{i=1}^M$ be the $M$ targets' predictions and $\{z_{t}^j\}_{j=1}^N$ ($f_0(z_t^j)\ge0.5$) be the $N$ person observations.

As shown in Fig. \ref{fig:association}(a), a conventional association matrix can be constructed, with each element reflecting the pairwise relationship between prediction and observation. The association matrix is equivalent to a bipartite graph, with the predictions and observations as nodes and the matrix elements as edge weights. The association problem is thus can be solved to obtain matching pairs with lowest cost via graph optimization methods such as max-flow or Hungarian algorithms. In our context, the prediction with matched observation is considered successfully tracked. A prediction (observation) with no match is considered as lost (new target).
However, the aforementioned association matrix may easily run into the risk of generating uncorrect pairs of prediction and observation.

Therefore, we propose to construct a novel joint association matrix $\mathcal{C}$ that can bridge the joint association optimization with a standard assignment problem. In our formulation, as illustrated in Fig. \ref{fig:association}(b), the rows and columns both comprise predictions and observations, and thus predictions (observations) can assign not only to their counterparts but also explicitly to themselves. In this way, the joint association matrix can be divided into 4 blocks, and each has meaningful representation when its element is chosen (i.e., lost, tracked or new target).

To be specific, matrix $\mathcal{C}$ is defined below:

\begin{equation}\label{equ:eq5}
\mathcal{C} =\left(
\begin{array}{cc}
	\Lambda & \Upsilon \\
	\Upsilon^T & \Gamma
\end{array}
\right)
\end{equation}
where $\mathcal{C}$ is a $(M+N)\times(M+N)$ square matrix, with row and column indices representing $M$ predictions and $N$ new observations.
Matrix $\mathcal{C}$ is composed of 4 blocks, where an element chosen in the submatrix $\Lambda(M\times M)$, $\Upsilon(M\times N)$ and $\Gamma(N\times N)$ implies that the corresponding target's state is judged as ``Lost'', ``Tracked'' and ``New'', respectively. $\Upsilon^T$ denotes the transpose of $\Upsilon$.

A type of function $p_{*}(\cdot,\cdot)$, $*\in \{\Lambda,\Upsilon,\Gamma\}$ is introduced to measure the pairwise relationship. A larger value of $p_{*}(\cdot,\cdot)$ indicates stronger correlation.

In block $\Lambda$, we define its element as follows:

\begin{equation}\label{equ:eq7}
\Lambda_{ij}=\begin{cases}
  p_{\Lambda}(x_t^i,x_t^j), & \text{if}~~i=j\\
  -\infty, & \text{otherwise}\\
\end{cases}
\end{equation}

Here, when a prediction is highly self-associated, we consider it to be lost. For two predictions of different person targets, we do not assign any coupling evidence and set the value to be $-\infty$.

In block $\Upsilon$, we define its element as follows:

\begin{equation}\label{equ:eq7}
\Upsilon_{ij}=  p_{\Upsilon}(x_t^i,z_t^j)
\end{equation}
where $i\in\{1, ..., M\}$ and $j\in\{1, ..., N\}$. The element definition indicates that a target is  successfully tracked when it is highly coupled with a person observation.

In block $\Gamma$, we define its element as follows:

\begin{equation}\label{equ:eq7}
\Gamma_{ij}=\begin{cases}
  p_{\Gamma}(z_t^i,z_t^j), & \text{if}~~i=j\\
  -\infty, & \text{otherwise}\\
\end{cases}
\end{equation}
Similar to the definition of the elements in $\Lambda$, a person observation that highly associates itself is considered as a new target. We also do not assign any coupling evidence between any two person observations and set the corresponding value to be $-\infty$.

The essential issue is how to define the functions $p_{*}(\cdot,\cdot)$ so that the aforementioned requirements can be satisfied. Many criteria based on multiple cues in the literature, such as appearance and motion, can be exploited. In this paper, we propose to use measurements tightly associated with our MBN network.
We define  $p_{*}(\cdot,\cdot)$ as the sum of two terms:

\begin{equation}\label{equ:p_sum}
p_{*}(\cdot,\cdot)=\lambda_{*}G_{*}(\cdot,\cdot)+(1-\lambda_{*})B_{*}(\cdot,\cdot),~~*\in \{\Lambda,\Upsilon,\Gamma\}
\end{equation}
where $G_{*}$ and $B_{*}$ are related to the confidence and location outputs of the MBN network, respectively.
The three parameters $\lambda_{*}$, $*\in \{\Lambda,\Upsilon,\Gamma\}$ are preset constants.

In particular, we define

\begin{equation}\label{equ:eq3}
G_{\Upsilon}(x_t^i,z_t^j)= f_i(z^j_t),\quad B_{\Upsilon} = IoU(x_t^i,z_t^j)
\end{equation}
where $f_i(z^j_t)$ denotes the output confidence by feeding observation $z^j_t$ into the $i$th instance detector. $IoU(x_t^i,z_t^j)$ is an intersection-over-union function which returns the area ratios of intersection and union between the bounding boxes of $x_t^i$ and $z_t^j$.

Then the terms $G_{\Lambda}$, $B_{\Lambda}$, $G_{\Gamma}$ and $B_{\Gamma}$ are defined as follows:

\begin{equation}\label{equ:lost}
\begin{aligned}
&G_{\Lambda}(x_t^i,x_t^i)= 1- f_i(x_t^i),\\
&B_{\Lambda}(x_t^i,x_t^i) = 1- \max\nolimits_{k=1}^{N} IoU(x_t^i,z_t^k)
\end{aligned}
\end{equation}
\begin{equation}\label{equ:new}
\begin{aligned}
&G_{\Gamma}(z_t^j,z_t^j)= 1- \max\nolimits_{k=1}^{M}f_k(z_t^j),\\
&B_{\Gamma}(z_t^j,z_t^j) = 1- \max\nolimits_{k=1}^{M} IoU(x_t^k,z_t^j)
\end{aligned}
\end{equation}
Specifically, Eq. (\ref{equ:lost}) indicates that a target is considered self-associated (or lost) when its own instance-subnet outputs low confidence and the predicted location is weakly coupled with the observations. Likewise, a person observation is considered self-associated (or new object), as implied by Eq. (\ref{equ:new}), when it retrieves low evidence from all available instance-subnets and their predicted locations.
We note that the terms $G_{*}$ and $B_{*}$, $*\in \{\Lambda,\Upsilon,\Gamma\}$ are all in the range $[0,\,1]$.

\subsection{Joint state inference} \label{sec:inference}

By constructing the joint association matrix, the joint tracking inference problem of all targets can be converted to an assignment problem by finding an optimal permutation vector $\bf{y}$ consisting of $\{1,2,...,M+N\}$. The energy function is formulated as:

\begin{equation}\label{equ:energy}
		{\bf{y}}^{*} = \arg\max_{\bf{y}} \sum_{k=1}^{M+N} \mathcal{C}(k,y^{k})
\end{equation}
where $y^{k}\in\{1,2,...,M+N\}$ is the $k$th element of $\bf{y}$ and $\mathcal{C}(k,y^{k})$ denotes the matrix element in row $k$ and column $y^{k}$ of $\mathcal{C}$.
Let $c_m$ to be the maximum element of $\mathcal{C}$, and replace each element $\mathcal{C}(i,j)$ with $c_m-\mathcal{C}(i,j)$ to obtain the matrix $\mathcal{C}'$. Then Eq. (\ref{equ:energy}) is equivalent to

\begin{equation}\label{equ:energy1}
		{\bf{y}}^{*} = \arg\min_{\bf{y}} \sum_{k=1}^{M+N} \mathcal{C}'(k,y^{k})
\end{equation}
We solve this energy function efficiently via the Hungarian algorithm \cite{munkres1957algorithms}.

We will update the instance-subnet $F_i$ when the target $i$ is in ``Tracked'' state but with $f_i(x_t^i)<\gamma$. For a person observation that is inferred as ``New'', a corresponding branch subnet will be initialized for it.
For a target $i$ judged in ``Lost'' state, if it has been in this state for $\tau$ consecutive frames, it will be transferred to the ``Discarded'' state. Otherwise it will continue to be predicted and participate in the joint inference in next frame.

{\bf Algorithm \ref{mot_alg}} depicts the procedure of the proposed INARLA framework.

\begin{algorithm}[!hbt]
    \caption{The overall procedure of our INARLA framework}\label{mot_alg}
    \begin{algorithmic}[1]
    \REQUIRE ~~\\
	A video sequence $\mathcal{V}$\\
	Initial MBN: Backbone-subnet $F_{bb}$ and det-pruning-subnet $F_{0}$
	
    \ENSURE ~~\\
	 Trajectories of targets $\mathcal{T}$\\

	\STATE Initialization: $\mathcal{T} \leftarrow \emptyset $
    \FOR {each frame $t$ in $\mathcal{V}$}
    \STATE Take the person proposals $\{z_j\}$ for a public human detector
	\STATE Use $F_{0}$ to reject false detections from $\{z_j\}$

	\FOR {each maintained person $i$ in $\mathcal{T}$}
		\STATE $F_{i}$ produces the predicted score and location (refer to Sec. \ref{sec:prediction})
	\ENDFOR
	\STATE  Construct association matrix and infer the state of each target  (refer to Sec. \ref{sec_states})
	\STATE Perform trajectory update of ``Tracked'' targets and initialization of ``New'' targets
	\STATE Update the MBN according to the state of each target
    \ENDFOR
    \end{algorithmic}
\end{algorithm}

\subsection{Assumption validation} \label{sec:Validation}

There exists a key assumption of selection in $\mathcal{C}$. That is, we have to ensure that once the elements in $\Upsilon$ are chosen,  the symmetric elements in $\Upsilon^T$ must be chosen as well, because we incorporate both predictions and observations in rows and columns and thus a matched pair should take two symmetric elements simultaneously.  Fortunately, due to the special structure of $\mathcal{C}$,  this assumption can be validated.

Let us take the joint matrix in Fig. \ref{fig:association}(b) for explanation. It can be observed that elements marked in red form a potential optimal solution, with each occupying distinct row and column and the elements being symmetric. However, the two elements marked in green in the left and the three elements marked in red in the right also seem to form a plausible optimal solution. But we will show that this is not true in our formulation context. Assume such asymmetric solution to be optimal. Let $A_{\Upsilon}$ be the sum of elements chosen in $\Upsilon$ and $A'_{\Upsilon}$ be the sum of elements chosen in $\Upsilon^T$.
If $A_{\Upsilon}>A'_{\Upsilon}$, it is obvious that we can choose the elements in $\Upsilon^T$ that are symmetric to those chosen in $\Upsilon$ to get a better solution. It conflicts with the optimum assumption. It is a similar case when $A_{\Upsilon}<A'_{\Upsilon}$.
It is almost impossible that $A_{\Upsilon}=A'_{\Upsilon}$ because we set matrix elements in floating numbers. In the extreme situation that $A_{\Upsilon}=A'_{\Upsilon}$, the problem has multiple optimal solutions even not expressed in our joint matrix.
In practice, extensive experimental results show that the optimal solution is symmetric.

\section{Experiments}\label{sec_experiment}

\subsection{Experimental settings} \label{sec:setting}

\noindent{\bf Dataset}\quad
The proposed method is evaluated on the 2D MOT 2015 benchmark dataset \cite{leal2015motchallenge}, which contains 11 sequences for training and 11 sequences for testing, consisting of sequences filmed by both static and moving cameras in unconstrained environments. The MOT benchmark releases ground truth for the training sequences. The human detection results provided by the benchmark dataset, which were generated by the ACF detector \cite{dollar2014fast}, are used in our evaluation so as to provide fair comparison with other MPT methods.

\noindent{\bf Evaluation metrics}\quad
Multiple metrics are used to evaluate the tracking performance as suggested
by the MOT research community \cite{bernardin2008evaluating,RistaniSZCT16eccv}, including Multiple Object Tracking Accuracy (MOTA, taking FN, FP and IDS into account), ID F1 Score (IDF1, the ratio of correctly identified detections over the average number of ground-truth and computed detections), Mostly Tracked targets (MT, the ratio of ground-truth trajectories that are covered by a track hypothesis for at least 80\% of their respective life span),  Mostly Lost targets (ML, The ratio of ground-truth trajectories that are covered by a track hypothesis for at most 20\% of their respective life span),  the total number of False Positives (FP),  the total number of False Negatives (FN),  the total number of ID Switches (IDS),  the total number of times a trajectory is Fragmented (Frag), and processing speed (Hz, in frames per second excluding the detector) on the benchmark.

\noindent{\bf MBN architecture}\quad
As mentioned in Sect. \ref{sec:MBN}, the structure of the backbone-subnet is the same as the lower five layers of CaffeNet used in Fast R-CNN \cite{girshick2015fast}. Specifically, the five convolutional layers have 96 kernels of size $11\times 11$, 256 kernels of size $5\times 5$, 384 kernels of size $3\times 3$, 384 kernels of size $3\times 3$ and 256 kernels of size $3\times 3$, respectively. The output feature maps of the first two convolutional layers are max-pooled ($3\times3$ kernel) and normalized before being fed into the next layer. Moreover, outputs of all the five layers are immediately filtered by a rectified linear unit (ReLU) before any pooling or normalization operation.
Branch subnets, including the det-pruning-subnet and instance-subnets, have the same structure, consisting of  a ROI layer, and three fully connected layers with size of 256, 256 and 2, respectively.

\noindent{\bf Implementation details}\quad
Our algorithm is implemented in python using Caffe platform.  The network $f_0$ (backbone-subnet with pruning subnet) is trained on the training set from \cite{leal2015motchallenge} for 40K SGD iterations and the learning rate is lowered by 0.1$\times$ in the last 10k iterations. We double the learning rate for training instance network for fast adaption and run for 50 iterations. The images on both training and testing phases are rescaled so that the shorter side of them is 600 pixels.
We set $\lambda_{\Lambda}=0.2$, $\lambda_{\Upsilon}=0.85$, $\lambda_{\Upsilon}=0.4$, $\sigma_s=(25,25,0.01)$, $\alpha=0.75$, $\gamma=0.5$ and $\tau=10$ in the experiments by empirical study. We will further discuss important parameter settings in ablation study (Sect. \ref{sec:ablation}).
Our algorithm runs on a PC with  8 cores of 3.70 GHZ CPU, and a Tesla K40 GPU.

\begin{table*}[!htb]
    \footnotesize
    \centering
    \caption{Quantitative evaluation results on the 2D MOT 2015 benchmark.}\label{tab_comparasion}
    \begin{tabular}{|@{\,}l@{\,}||@{\,}c@{\,}|@{\,}c@{\,}|@{\,}c@{\,}|@{\,}c@{\,}|@{\,}c@{\,}|@{\,}c@{\,}|@{\,}c@{\,}|@{\,}c@{\,}|@{\,}c@{\,}|}
	\hline
    \textbf{Algorithm} & \textbf{MOTA(\%)}\,$\uparrow$\, & \textbf{IDF1(\%)}\,$\uparrow$\, & \textbf{MT(\%)}\,$\uparrow$\,  & \textbf{ML(\%)}\,$\downarrow$\, & \textbf{FP}\,$\downarrow$\, & \textbf{FN}\,$\downarrow$\, & \textbf{IDS}\,$\downarrow$\, & \textbf{Frag}\,$\downarrow$\, & \textbf{Hz}\,$\uparrow$\, \\		
    \hline
	\hline
	{SiameseCNN (2016)\cite{Leal-TaixeCS16cvpr}} $\dag$ & 29.0 & 34.3 & 8.5 & 48.4 & 5160 & 37798 & 639 & 1316 & 52.8 \\
	\hline
	{CNNTCM (2016)\cite{WangWSZLCW16cvpr}} $\dag$& 29.6 & 36.8 & 11.2 & {44.0} & 7786 & {34733} & 712 & \textbf{943} & 1.7	 \\
	\hline
	{QuadMOT (2017)\cite{SonBCH17cvpr}} $\dag$& 33.8 & 40.4 & 12.9 & {36.9} & 7898 & {32061} & 703 & 1430 & 3.7 \\
    \hline
	\hline
	{TSDA\_OAL (2017)\cite{JuIET2017} } & 18.6 & 36.1  &9.4  & 42.3  &16350  & 32853 & 806 &  1544  & 19.7  \\
	\hline
	{RNN\_LSTM (2016)\cite{MilanRDSR16}}  & 19.0 & 17.1 &5.5 & 45.6 & 11578 & 36706 & 1490 & 2081 & \textbf{165.2}\\
	\hline
	{OMT\_DFH (2017)\cite{ju2017online} } & 21.2 & 37.3 & 7.1  & 46.5 & 13218 & 34657 & 563  & {1255} &	 {28.6}  \\
	\hline
	{EAMTTpub (2016)\cite{eccvMatilla16}} & 22.3 & 32.8 & 5.4  & 52.7 & 7924 & 38982 & 833  &	1485  &	12.2  \\
	\hline
	{oICF (2016)\cite{avssKieritzBHA16}} & 27.1 & 40.5 & 6.4  & 48.7 &7594 & 36757 & {454} & 1660 & 1.4 \\
	\hline
	{SCEA (2016)\cite{yoon2016online}} & 29.1 & 37.2  & 8.9 & 47.3 &{6060} & 36912 & 604 & {1182} & 6.8 \\
	\hline
	{MDP (2015)\cite{xiang2015learning}} &30.3 & {44.7} & \textbf{13.0} & 38.4 & 9717 & 32422 & 680 & 1500 & 1.1 \\	
	\hline
	{DCCRF (2018)\cite{Zhou2018CSVT}} & {33.6} & 39.1 & 10.4 & 37.6 & 5917 & 34002 & 866 & 1566 & 0.1 \\
	\hline
	{AM (2017)\cite{ChuOLWLY17iccv}} & {34.3} & \textbf{48.3} & 11.4 & 43.4 & \textbf{5154} & 34848 & \textbf{348} & 1463 & 0.5 \\
	\hline
    {\textbf{INARLA (Ours)}} & \textbf{34.7} & 42.1 &	{12.5} &	\textbf{30.0} &	9855 & \textbf{29158} &	1112 &	2848 &	 2.6	\\
	\hline
	\end{tabular}\vspace{0.2ex}\\
\raggedright
 \begin{tabular}{l} \hspace{13ex} $\dag$ denotes offline methods. \end{tabular}
\end{table*}

\begin{table*}[!hbt]
    \footnotesize
    \centering
    \caption{Object density (OPF) and tracking efficiency (FPS) of each sequence on test set.}\label{tab_fps}
    \begin{tabular}{|l||@{\hspace{1ex}}c@{\hspace{1ex}}|@{\hspace{1ex}}c@{\hspace{1ex}}||l||@{\hspace{1ex}}c@{\hspace{1ex}}|@{\hspace{1ex}}c@{\hspace{1ex}}|}
	\hline
    \textbf{Sequence} & \textbf{Density} & \textbf{Speed} & \textbf{Sequence} & \textbf{Density} & \textbf{Speed}\\	
    \hline
    \emph{ETH-Crossing}  &4.6      &6.1  & \emph{ETH-Jelmoli}  &5.8    & 4.6   \\
    \hline
	\emph{ETH-Linthescher}  &7.5      &5.4  & \emph{KITTI-19}  &5    & 4.2 \\
		\hline
	\emph{TUD-Crossing}  &5.5      &4.8   & \emph{KITTI-16}  &8.1    & 3.2\\
    \hline
    \emph{ADL-Rundle-3}  &16.3      &2.0 & \emph{Venice-1}  &10.1      & 3.3\\
		\hline
		\emph{ADL-Rundle-1}  &18.6    & 1.3   & \emph{PETS09-S2L2}  &22.1    & 1.0\\
    \hline
    \emph{AVG-TownCentre}  &15.9    & 1.0 & & &\\
		\hline
	\end{tabular}
\end{table*}

\subsection{Benchmark evaluation}
We compare our INARLA tracker with nine recent online MPT methods that published their results on the 2D MOT 2015 benchmark, including
TSDA\_OAL \cite{JuIET2017}, RNN\_LSTM \cite{MilanRDSR16}, OMT\_DFH \cite{ju2017online}, EAMTTpub \cite{eccvMatilla16}, oICF \cite{avssKieritzBHA16}, SCEA \cite{yoon2016online}, MDP \cite{xiang2015learning}, DCCRF \cite{Zhou2018CSVT} and AM \cite{ChuOLWLY17iccv}. Among them, RNN\_LSTM, DCCRF and AM are deep learning-based methods.
We also include three recent deep learning-based offline MPT methods (i.e., SiameseCNN \cite{Leal-TaixeCS16cvpr}, CNNTCM \cite{WangWSZLCW16cvpr} and QuadMOT \cite{SonBCH17cvpr}) for comparison.
Table \ref{tab_comparasion} summarizes the quantitative comparison results, and the best result in each metric is marked in bold font.
The up-arrow next to a metric indicates higher values are better, while the down-arrow indicates lower values are better.

\begin{figure*}[!htb]
    \centering
    \includegraphics[width=0.8\linewidth]{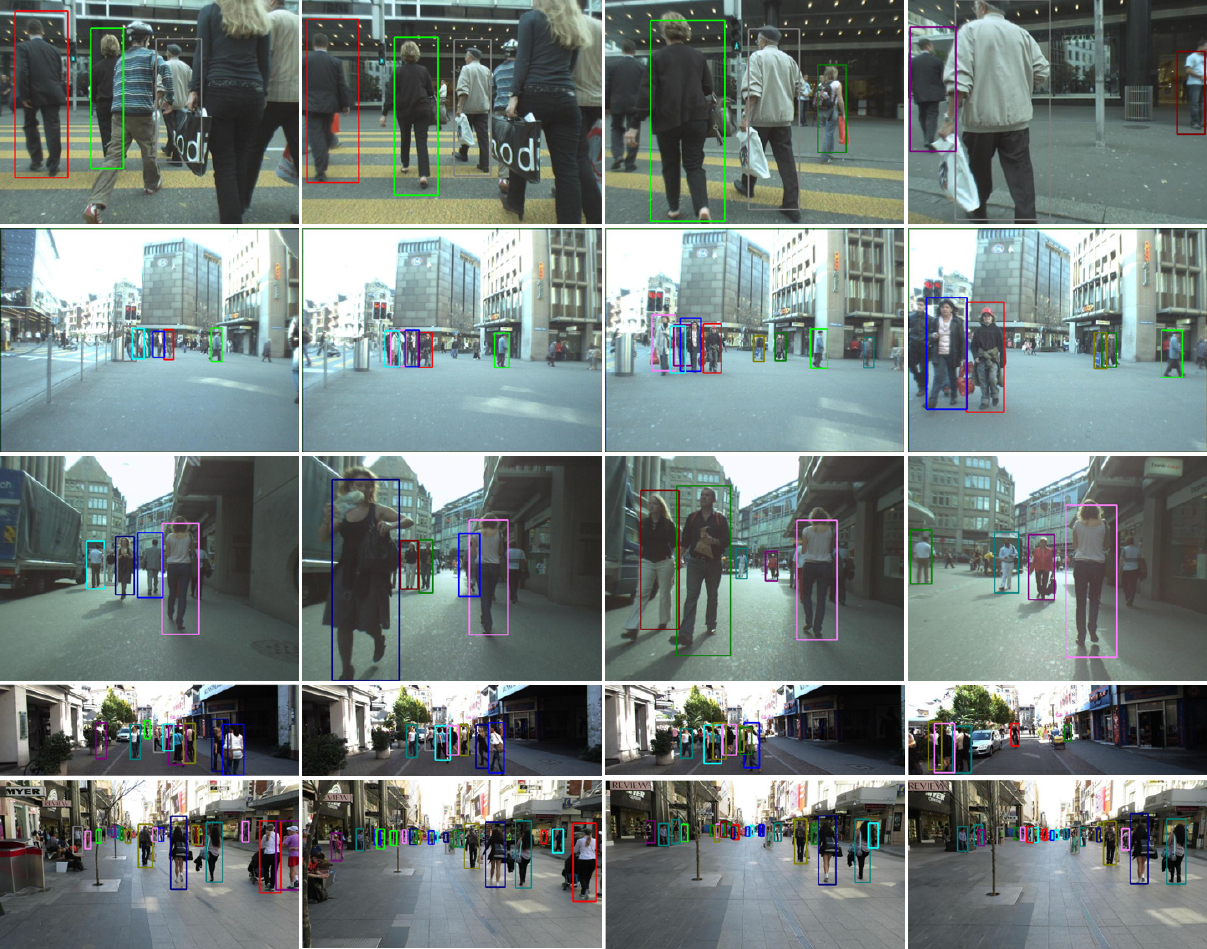}
    \caption{Our tracking results on representative MOTChallenge dynamic scenes including ETH-Crossing, ETH-Jelmoli, ETH-Linthescher, KITTI-19 and ADL-Rundle-1, from top to bottom. }\label{fig:dynamic}

\end{figure*}

Among these metrics, MOTA is an integrated metric that summarizes multiple aspects of tracking performance and is used by the MOT benchmark for ranking the trackers. Our method achieves the highest MOTA against these recent methods including the deep learning-based methods.
Moreover, our method also achieves the best performance in terms of ML and FN since our network achieves robust performance in the presence of missing detections.
The outstanding performance demonstrates the advantages of our MBN network and joint state inference solver.
However, working in a frame-by-frame way, our method will recover targets judged as ``Lost'' for many times, resulting in a high Frag value. This can be further addressed by introducing a proper post-processing strategy.
Fig. \ref{fig:static} and Fig.~\ref{fig:dynamic} illustrate  our tracking results on the test set of the MOT benchmark in static and dynamic scenes, respectively.

Our algorithm runs at around 2.6 frames per second without code optimization. Note that the number of tracked objects actually affects the running speed. Therefore, we show in Table \ref{tab_fps} the relationship between the density (objects per frame, OPF) and the processing speed (frames per second, FPS) on each sequence of the test set. It can be inferred from Table \ref{tab_fps} that the speed of a single instance tracker roughly ranges from 20 to 30 fps. Due to the properties of our MBN, we are confident that improved processing efficiency can be achieved by parallel implementation in branch subnets.

\subsection{Ablation study}\label{sec:ablation}

The contributions of different components in our method
are assessed on the 2D MOT 2015 benchmark. The ablation study is conducted on the training set because the annotations of the test set are not released and the benchmark webpage limits evaluation submissions (a user can only post a submission every three days and submit no more than 3 times in total).
The 11 training sequences are partitioned into training and validation subsets to
analyze the proposed algorithm, with 5 sequences (TUD-Stadtmitte, ETH-Bahnhof, ADL-Rundle-8, PETS09-S2L1, KITTI-13) for training and the rest for validation.

Table \ref{tab:ablation} reports the quantitative evaluation results of different versions of our MPT method in ablation study. The results of the full version of our method, which contains all the proposed components, are shown in the last row of the table. Below we evaluate and analyze each component of the proposed MPT method in detail.

\begin{table*}[!htb]
    \footnotesize
    \centering
    \caption{Quantitative comparison of different versions of our method in ablation study.}\label{tab:ablation}
    \begin{tabular}{|@{\,}l@{\,}||@{\,}c@{\,}|@{\,}c@{\,}|@{\,}c@{\,}|@{\,}c@{\,}|@{\,}c@{\,}|@{\,}c@{\,}|@{\,}c@{\,}|@{\,}c@{\,}|}
	\hline
    \textbf{Version} & \textbf{MOTA(\%)}\,$\uparrow$\, & \textbf{IDF1(\%)}\,$\uparrow$\, & \textbf{MT(\%)}\,$\uparrow$\,  & \textbf{ML(\%)}\,$\downarrow$\, & \textbf{FP}\,$\downarrow$\, & \textbf{FN}\,$\downarrow$\, & \textbf{IDS}\,$\downarrow$\, & \textbf{Frag}\,$\downarrow$\, \\		
	\hline
	\hline
                    no\_aux\_loss & 40.2&  48.3& 20.9& 33.9&   3286&  9233&   223&   510   \\
         no\_pruning & 32.2&  29.9& 17.0& 35.2&   3549& 10286&   605&   779   \\
                no\_update & 38.7&  45.5& 18.3& 34.8&   3451&  9404&   216&   488   \\
    \hline
	\hline
             only\_IoU & 38.5&  44.1& 19.6& 36.5&   3164&  9711&   237&   458   \\
           only\_confidence & 36.4&  43.1& 16.5& 39.1&   3220& 10066&   267&   464   \\
             balance\_learned & 39.7&  47.6& 20.4& 35.7&   3370&  9258&   219&   467   \\
         greedy & 25.4&  31.8& 17.4& 37.8&   5579&  9726&   597&   753   \\
	\hline
	\hline
                 with vgg16 & 39.0&  46.7& 19.1& 36.1&   3058&  9708&   242&   489   \\
              with vgg\_m & 40.6&  47.3& 19.1& 36.5&   3031&  9397&   237&   487   \\
	\hline
	\hline
                     full & 41.1&  48.7& 21.7& 35.7&   3097&  9248&   201&   461   \\
	\hline
	\end{tabular}
\end{table*}

{\it 1) MBN network:} The offline training of our MBN network is augmented with an auxiliary subnet in a multi-task optimization scheme, as described in Sect. \ref{sec:network_learn}. And it aims to make the MBN network more discriminative for our MPT task. To evaluate its effectiveness, we remove the auxiliary subnet and set $\mu=0$ in Eq. (\ref{eq:multi}) for offline model training, and this version of our method is termed ``no\_aux\_loss''. From Table \ref{tab:ablation}, we can observe that its MOTA performance drops by about 1\% with most of the other metrics also degraded. The increase in FP reveals it includes more false human detections. These results demonstrate the positive role of the auxiliary subnet.

The ``no\_pruning'' version of our method denotes our framework does not include the process of the det-pruning-subnet that aims to filter out false human detections. As can be observed, its MOTA drops dramatically to 32.2\%, with a decrease of 8.9\%. The FP metric increases from 3097 to 3451. A sharp performance degradation can be viewed in most of the metrics, which demonstrates the significant effectiveness of the det-pruning-subnet.

The instance-subnets of our MBN network are dynamically added and trained online. They are also updated during tracking so as to adapt to appearance changes of corresponding human instances. The ``no\_update'' version denotes an instance-subnet will not be updated after it is trained. As shown in Table \ref{tab:ablation}, the deterioration in all the metrics except ML reveals the importance of online update.

{\it 2) Association matrix:} The second group of rows in Table \ref{tab:ablation} evaluates the effectiveness of our data association component that builds upon the constructed association matrix. As depicted by Eq. (\ref{equ:p_sum}), elements of the association matrix involves two terms (i.e., output confidence and IoU) and three parameters (i.e., $\lambda_{\Lambda}$, $\lambda_{\Upsilon}$ and $\lambda_{\Gamma}$). We carry out experiments to evaluate their influence on our method's performance.

The ``only\_confidence'' and ``only\_IoU'' versions of our method denote Eq. (\ref{equ:p_sum}) only contains the confidence- or IoU-related term, corresponding to setting $\lambda_{*}=1$ and $\lambda_{*}=0$ ($*\in \{\Lambda,\Upsilon,\Gamma\}$), respectively.
Performance degradation in all the metrics are witnessed from Table \ref{tab:ablation} for both these two versions. We can also infer that the IoU-related term has a larger impact on our method's performance because ``only\_IoU'' performs better than ``only\_confidence'' in
the evaluation metrics.

We further discuss the problem of balancing the two terms in Eq. (\ref{equ:p_sum}), i.e., choosing the best values for parameters $\lambda_{*}$ ($*\in \{\Lambda,\Upsilon,\Gamma\}$). A balance-learning scheme was tried to find the optimal parameter setting. The scheme is designed as follows. Given initial parameter setting of $\lambda_{*}$, the proposed algorithm is run on the training set. Then we check the ground-truth for a pair in function $p_{*}(\cdot,\cdot)$ every frame, and the expected output of $p_{*}(\cdot,\cdot)$ is set as 1 if the pair is matched and 0 otherwise. We learn $\lambda_{*}$ by minimizing the sum of squared errors of actual and expected outputs. The process is executed for several iterations with the learned value of $\lambda_{*}$ as new initial setting. The best results of this balance-learning scheme are shown in Table \ref{tab:ablation} as ``balance\_learned''.
As can be seen, this scheme does not work quite well. It performs worse than the ``full'' version in which $\lambda_{*}$ are manually set by empirical study. In future, we will try new schemes to handle this problem.

To further analyze the contribution of our association component, we replace it with a simple greedy association algorithm. That is, in the association stage, a new person observation will be assigned to a tracked target who has the largest IoU ratio of bounding boxes with it. This version of our method is termed ``greedy''. As exhibited in Table \ref{tab:ablation}, its performance worsens sharply in all the metrics, which instead reveals the significant role of the proposed association component.

{\it 3) Choices of backbone-subnet:} As described in Sect. \ref{sec:setting}, the backbone-subnet of our MBN network is CaffeNet, a small-scale neural network. Here we make other choices for the backbone-subnet to evaluate their impact on the performance. Specifically, we use vgg\_cnn\_m\_1024 \cite{ChatfieldSVZ14bmvc} and vgg16 \cite{SimonyanZ14acorr} network models as the backbone-subnet. The vgg\_cnn\_m\_1024 model is the same deep as CaffeNet but is wider, and the vgg16 model is very deep with 16 layers. With these two models, the corresponding versions of our method are termed ``with vgg\_m'' and ``with vgg16'' in Table \ref{tab:ablation}. It can be observed that ``with vgg\_m'' has almost the same performance with ``full'', with 0.5\% decrease in MOTA. However, ``with vgg16'' shows larger degradation in performance with MOTA decreased by 2.1\%. We visualized the tracking results and took in-depth analysis, and found that the ``with vgg16'' version did not work well on small-sized persons.  It may be attributed to that a small image region contains less appearance details that are important for discriminating instances of the same category (e.g. human) and the feature extracted by the deep vgg16 model is less reflective of those details since the vgg16 architecture induces stronger reduction of subtle features (e.g., with more max-pooling layers than CaffeNet), as also reported in previous work \cite{Wang2015Visual,LiWLL18jcam}.  It is also worth noting that the ``full'', ``with vgg\_m'' and ``with vgg16'' versions run at about 2.7, 2.2, 1.6 FPS averagely on the validation set, respectively. The foregoing comparison reveals the ``full'' version performs the best in both accuracy and efficiency among the three versions.

\begin{figure}[!htb]
    \centering
    \includegraphics[width=0.93 \linewidth]{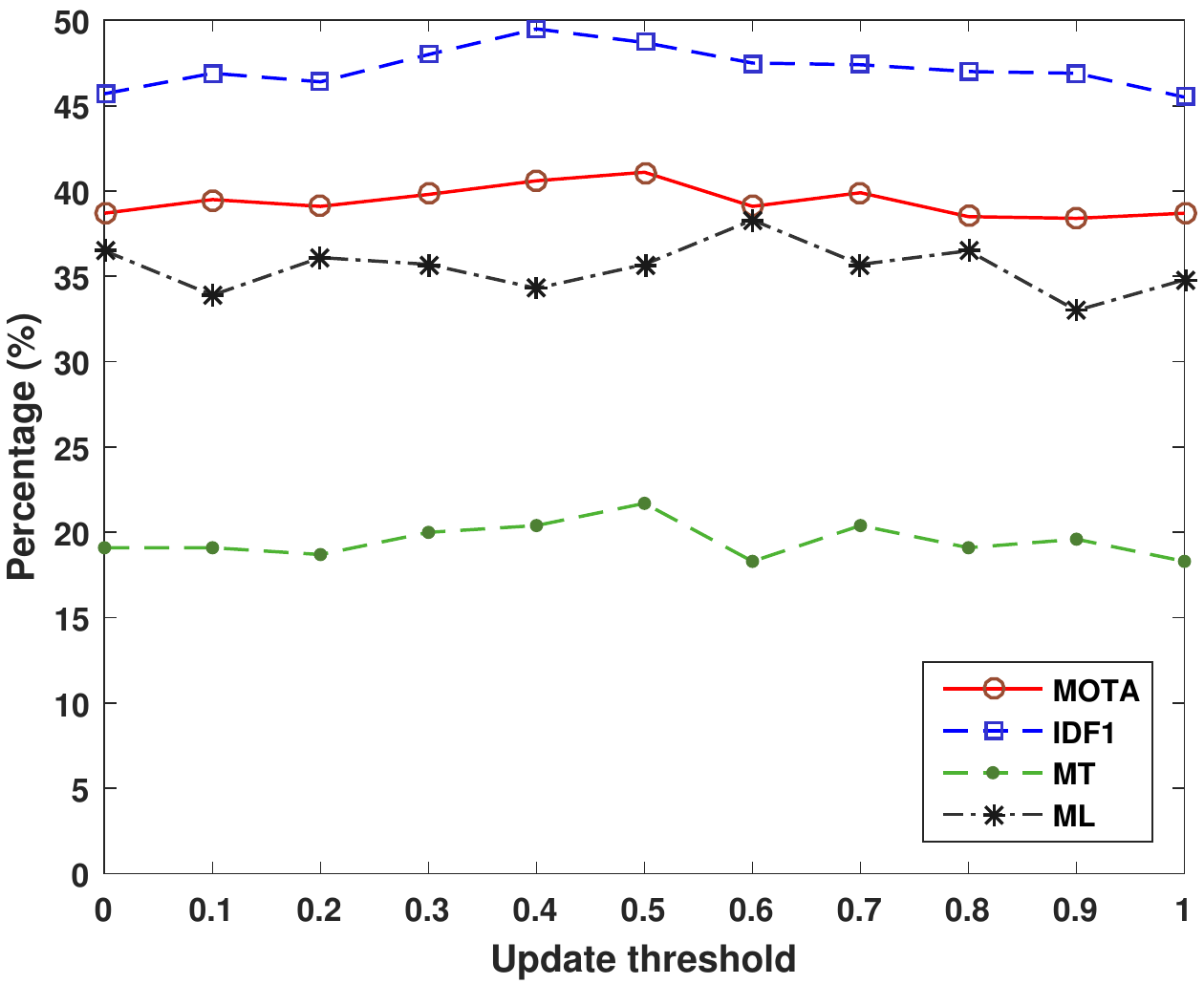}
    \caption{Analysis of update threshold on the validation set.}\label{fig:updateThr}

\end{figure}

{\it 4) Update threshold:} To evaluate the influence of the update threshold $\gamma$ on our method's performance, we change its value while fixing the values of the other parameters. The results are plotted in Fig. \ref{fig:updateThr} with four metrics (MOTA, IDF1, MT and ML) that are expressed in percentage. As can be observed, the performance is the best when the update threshold is at around 0.5, but the performance does not exhibit a sharp change as the threshold changes.

\section{Conclusion}\label{sec_conclusion}
In this paper, we have introduced a novel deep learning based online multi-person tracking approach that emphasizes instance-aware representation learning with the MBN network. While the backbone-subnet provides robust deeply-learned image feature, the instance-subnets cast instance-level appearance discrimination to reduce ambiguities between different targets and release the burden of later data association. We construct an association matrix based on the outputs of the MBN network for joint state inference of the targets, where a simple yet effective solver is developed thanks to the powerful support from MBN. The effectiveness of our approach is verified through extensive experimental evaluation with recent MPT methods.

There are several directions that we can improve the proposed INARLA framework in future. First, the backbone-subnet of our MBN network will be enhanced to empower its extracted feature more robustness and discrimination.
Our approach can handle small-sized objects better by making the feature extraction process adapt to different sizes of objects.
Second, a more efficient model should be designed for the instance-subnet. This is because we found in experiments that online training and updating of instance-subnets often occupy more than half of the total processing time although the instance-subnet in our MBN network has a light-weight structure.
Recent works show that correlation filter models can achieve good accuracy at high running speed in single object tracking.
We will make in-depth attempts to incorporate such models into our MBN network since they also involve convolution.
Third, more effort will be devoted to the state inference procedure. We will investigate more effective terms for  composing elements of the association matrix and exploit new data association algorithms for the online MPT task.
Moreover, we intend to extend our work to incorporate full category detection and form a unified framework.

\section*{Acknowledgement}
{\small
This work was supported by the National Natural Science Foundation of China (61876045, U1811463), Zhujiang Science and Technology New Star Project of Guangzhou (201906010057), the Major Program of Science and Technology Planning Project of Guangdong Province (2017B010116003), and Guangdong Natural Science Foundation (2016A030313285). The authors would like to thank Shiyi Hu and Xu Cai who partly joined this work when they were graduate students at Sun Yat-sen University.
}

\section*{References}


{\small
\bibliographystyle{elsarticle-num}
\bibliography{MPT}

\begin{thebibliography}{10}
\expandafter\ifx\csname url\endcsname\relax
  \def\url#1{\texttt{#1}}\fi
\expandafter\ifx\csname urlprefix\endcsname\relax\def\urlprefix{URL }\fi
\expandafter\ifx\csname href\endcsname\relax
  \def\href#1#2{#2} \def\path#1{#1}\fi

\bibitem{LinWZF016pami}
L.~Lin, G.~Wang, W.~Zuo, X.~Feng, L.~Zhang, Cross-domain visual matching via
  generalized similarity measure and feature learning, TPAMI 39~(6) (2016)
  1089--1102.

\bibitem{XieDZWF17pami}
J.~Xie, G.~Dai, F.~Zhu, E.~K. Wong, Y.~Fang, Deepshape: Deep-learned shape
  descriptor for 3d shape retrieval, TPAMI 39~(7) (2017) 1335--1345.

\bibitem{WuYL17pr}
Y.~Wu, F.~Yin, C.~Liu, Improving handwritten chinese text recognition using
  neural network language models and convolutional neural network shape models,
  Pattern Recognition 65 (2017) 251--264.

\bibitem{milan2015joint}
A.~Milan, L.~Leal-Taix{\'e}, K.~Schindler, I.~Reid, Joint tracking and
  segmentation of multiple targets, in: CVPR, 2015, pp. 5397--5406.

\bibitem{wang2015learning}
S.~Wang, C.~Fowlkes, Learning optimal parameters for multi-target tracking, in:
  BMVC, 2015.

\bibitem{munkres1957algorithms}
J.~Munkres, Algorithms for the assignment and transportation problems, Journal
  of the Society for Industrial and Applied Mathematics 5~(1) (1957) 32--38.

\bibitem{ChoiS10eccv}
W.~Choi, S.~Savarese, Multiple target tracking in world coordinate with single,
  minimally calibrated camera, in: ECCV, 2010, pp. 553--567.

\bibitem{DalalT05cvpr}
N.~Dalal, B.~Triggs, Histograms of oriented gradients for human detection, in:
  CVPR, 2005, pp. 886--893.

\bibitem{Andriyenko2011Multi}
A.~Andriyenko, K.~Schindler, Multi-target tracking by continuous energy
  minimization, in: CVPR, 2011, pp. 1265--1272.

\bibitem{QianYG13pr}
J.~Qian, J.~Yang, G.~Gao, Discriminative histograms of local dominant
  orientation {(D-HLDO)} for biometric image feature extraction, Pattern
  Recognition 46~(10) (2013) 2724--2739.

\bibitem{Bae2014Robust}
S.~H. Bae, K.~J. Yoon, Robust online multi-object tracking based on tracklet
  confidence and online discriminative appearance learning, in: CVPR, 2014, pp.
  1218--1225.

\bibitem{BenfoldR11cvpr}
B.~Benfold, I.~D. Reid, Stable multi-target tracking in real-time surveillance
  video, in: CVPR, 2011, pp. 3457--3464.

\bibitem{LiWZLLW17tip}
H.~Li, H.~Wu, H.~Zhang, S.~Lin, X.~Luo, R.~Wang, Distortion-aware correlation
  tracking, {IEEE} TIP 26~(11) (2017) 5421--5434.

\bibitem{KimKFH12accv}
S.~Kim, S.~Kwak, J.~Feyereisl, B.~Han, Online multi-target tracking by large
  margin structured learning, in: ACCV, 2012, pp. 98--111.

\bibitem{lenz2015followme}
P.~Lenz, A.~Geiger, R.~Urtasun, et~al., Followme: Efficient online min-cost
  flow tracking with bounded memory and computation, in: ICCV, 2015, pp.
  4364--4372.

\bibitem{WuGCW18nca}
H.~Wu, C.~Gao, Y.~Cui, R.~Wang, Multipoint infrared laser-based detection and
  tracking for people counting, Neural Computing and Applications 29~(5) (2018)
  1405--1416.

\bibitem{FelzenszwalbGMR10pami}
P.~F. Felzenszwalb, R.~B. Girshick, D.~A. McAllester, D.~Ramanan, Object
  detection with discriminatively trained part-based models, TPAMI 32~(9)
  (2010) 1627--1645.

\bibitem{Wang2015Visual}
L.~Wang, W.~Ouyang, X.~Wang, H.~Lu, Visual tracking with fully convolutional
  networks, in: ICCV, 2015, pp. 3119--3127.

\bibitem{LiWLL18jcam}
H.~Li, H.~Wu, S.~Lin, X.~Luo, Coupling deep correlation filter and online
  discriminative learning for visual object tracking, Journal of Computational
  and Applied Mathematics 329 (2018) 191--201.

\bibitem{Cui2016CVPR}
Z.~Cui, S.~Xiao, J.~Feng, S.~Yan, Recurrently target-attending tracking, in:
  CVPR, 2016.

\bibitem{Ondruska2016Deep}
P.~Ondruska, I.~Posner, Deep tracking: Seeing beyond seeing using recurrent
  neural networks, in: AAAI, 2016, pp. 3361--3368.

\bibitem{Leal2016CVPRWorkshops}
L.~Leal-Taixe, C.~Canton-Ferrer, K.~Schindler, Learning by tracking: Siamese
  cnn for robust target association, in: CVPR Workshops, 2016.

\bibitem{gaidon2015online}
A.~Gaidon, E.~Vig, Online domain adaptation for multi-object tracking, in:
  BMVC, 2015, pp. 1--13.

\bibitem{yoon2016online}
J.~H. Yoon, C.-R. Lee, M.-H. Yang, K.-J. Yoon, Online multi-object tracking via
  structural constraint event aggregation, in: CVPR, 2016.

\bibitem{Tang2017Multiple}
S.~Tang, M.~Andriluka, B.~Andres, B.~Schiele, Multiple people tracking by
  lifted multicut and person re-identification, in: CVPR, 2017.

\bibitem{WangTFF16pami}
X.~Wang, E.~T{\"{u}}retken, F.~Fleuret, P.~Fua, Tracking interacting objects
  using intertwined flows, TPAMI 38~(11) (2016) 2312--2326.

\bibitem{LeibeSG07iccv}
B.~Leibe, K.~Schindler, L.~J.~V. Gool, Coupled detection and trajectory
  estimation for multi-object tracking, in: ICCV, 2007, pp. 1--8.

\bibitem{WangTFF14eccv}
X.~Wang, E.~T{\"{u}}retken, F.~Fleuret, P.~Fua, Tracking interacting objects
  optimally using integer programming, in: ECCV, 2014, pp. 17--32.

\bibitem{MaksaiWFF17iccv}
A.~Maksai, X.~Wang, F.~Fleuret, P.~Fua, Non-markovian globally consistent
  multi-object tracking, in: ICCV, 2017, pp. 2563--2573.

\bibitem{xiang2015learning}
Y.~Xiang, A.~Alahi, S.~Savarese, et~al., Learning to track: Online multi-object
  tracking by decision making, in: ICCV, 2015, pp. 4705--4713.

\bibitem{MilanRDSR16}
A.~Milan, S.~H. Rezatofighi, A.~R. Dick, K.~Schindler, I.~D. Reid, Online
  multi-target tracking using recurrent neural networks, ArXiv (2016)
  abs/1604.03635.

\bibitem{girshick2015fast}
R.~Girshick, Fast {R-CNN}, in: ICCV, 2015, pp. 1440--1448.

\bibitem{krizhevsky2012imagenet}
A.~Krizhevsky, I.~Sutskever, G.~E. Hinton, Imagenet classification with deep
  convolutional neural networks, in: NIPS, 2012, pp. 1097--1105.

\bibitem{YunRV14nips}
H.~Yun, P.~Raman, S.~V.~N. Vishwanathan, Ranking via robust binary
  classification, in: NIPS, 2014, pp. 2582--2590.

\bibitem{Shrivastava2016CVPR}
A.~Shrivastava, A.~Gupta, R.~Girshick, Training region-based object detectors
  with online hard example mining, in: CVPR, 2016.

\bibitem{leal2015motchallenge}
L.~Leal-Taix{\'e}, A.~Milan, I.~Reid, S.~Roth, K.~Schindler, Motchallenge 2015:
  Towards a benchmark for multi-target tracking, ArXiv (2015) abs/1504.01942.

\bibitem{dollar2014fast}
P.~Doll{\'a}r, R.~Appel, S.~Belongie, P.~Perona, Fast feature pyramids for
  object detection, TPAMI 36~(8) (2014) 1532--1545.

\bibitem{bernardin2008evaluating}
K.~Bernardin, R.~Stiefelhagen, Evaluating multiple object tracking performance:
  the {CLEAR} {MOT} metrics, EURASIP Journal on Image and Video Processing
  (2008) 1--10.

\bibitem{RistaniSZCT16eccv}
E.~Ristani, F.~Solera, R.~S. Zou, R.~Cucchiara, C.~Tomasi, Performance measures
  and a data set for multi-target, multi-camera tracking, in: ECCV Workshops,
  2016, pp. 17--35.

\bibitem{Leal-TaixeCS16cvpr}
L.~Leal{-}Taix{\'{e}}, C.~Canton{-}Ferrer, K.~Schindler, Learning by tracking:
  Siamese {CNN} for robust target association, in: CVPR Workshops, 2016.

\bibitem{WangWSZLCW16cvpr}
B.~Wang, L.~Wang, B.~Shuai, Z.~Zuo, T.~Liu, K.~L. Chan, G.~Wang, Joint learning
  of convolutional neural networks and temporally constrained metrics for
  tracklet association, in: CVPR Workshops, 2016.

\bibitem{SonBCH17cvpr}
J.~Son, M.~Baek, M.~Cho, B.~Han, Multi-object tracking with quadruplet
  convolutional neural networks, in: CVPR, 2017, pp. 3786--3795.

\bibitem{JuIET2017}
J.~Ju, D.~Kim, B.~Ku, D.~K. Han, H.~Ko, Online multi-person tracking with
  two-stage data association and online appearance model learning, IET Computer
  Vision 11 (2017) 87--95.

\bibitem{ju2017online}
J.~Ju, D.~Kim, B.~Ku, D.~K. Han, H.~Ko, Online multi-object tracking with
  efficient track drift and fragmentation handling, J. Opt. Soc. Am. A Opt.
  Image Sci. Vis. 34~(2) (2017) 280--293.

\bibitem{eccvMatilla16}
R.~Sanchez{-}Matilla, F.~Poiesi, A.~Cavallaro, Online multi-target tracking
  with strong and weak detections, in: ECCV Workshops, 2016, pp. 84--99.

\bibitem{avssKieritzBHA16}
H.~Kieritz, S.~Becker, W.~H{\"{u}}bner, M.~Arens, Online multi-person tracking
  using integral channel features, in: AVSS, 2016, pp. 122--130.

\bibitem{Zhou2018CSVT}
H.~Zhou, W.~Ouyang, J.~Cheng, X.~Wang, H.~Li, Deep continuous conditional
  random fields with asymmetric inter-object constraints for online
  multi-object tracking, IEEE TCSVT (2018) 1--12, online available.

\bibitem{ChuOLWLY17iccv}
Q.~Chu, W.~Ouyang, H.~Li, X.~Wang, B.~Liu, N.~Yu, Online multi-object tracking
  using {CNN}-based single object tracker with spatial-temporal attention
  mechanism, in: ICCV, 2017, pp. 4846--4855.

\bibitem{ChatfieldSVZ14bmvc}
K.~Chatfield, K.~Simonyan, A.~Vedaldi, A.~Zisserman, Return of the devil in the
  details: Delving deep into convolutional nets, in: BMVC, 2014.

\bibitem{SimonyanZ14acorr}
K.~Simonyan, A.~Zisserman, Very deep convolutional networks for large-scale
  image recognition, ArXiv (2014) abs/1409.1556.

\end{thebibliography}
}

%
%
%

\end{document}